%% file: main.tex
\documentclass{article}

\usepackage{amsmath, amsfonts}
\usepackage{microtype}
\usepackage{graphicx}
\usepackage{subfigure}
\usepackage{booktabs} 
\usepackage{mathtools}
\usepackage{bm}
\usepackage{caption}

\usepackage{hyperref}

\usepackage[accepted]{icml2021}

\DeclarePairedDelimiterX{\infdivx}[2]{(}{)}{%
  #1\;\delimsize\|\;#2%
}

\icmltitlerunning{Parallel and Flexible Sampling from Autoregressive Models}

\begin{document}

\twocolumn[
\icmltitle{Parallel and Flexible Sampling from Autoregressive\\ Models via Langevin Dynamics}



\icmlsetsymbol{equal}{*}

\begin{icmlauthorlist}
\icmlauthor{Vivek Jayaram}{equal,uw}
\icmlauthor{John Thickstun}{equal,uw}
\end{icmlauthorlist}

\icmlaffiliation{uw}{Department of Computer Science, University of Washington}

\icmlcorrespondingauthor{Vivek Jayaram}{vjayaram@cs.washington.edu}
\icmlcorrespondingauthor{John Thickstun}{thickstn@cs.washington.edu}

\icmlkeywords{Machine Learning, ICML}

\vskip 0.3in
]



\printAffiliationsAndNotice{\icmlEqualContribution} 

\input{src/abstract}
\input{src/intro}
\input{src/related}
\input{src/method}
\input{src/experiments}
\input{src/conclusion}

\section*{Acknowledgements}
We thank Zaid Harchaoui, Sham M. Kakade, Steven Seitz, and Ira Kemelmacher-Shlizerman for valuable discussion and computing resources. This work was supported by a Qualcomm Innovation Fellowship.

\bibliography{references}
\bibliographystyle{icml2021}

\clearpage
\appendix
\onecolumn

\input{src/appendix.tex}

\end{document}

%% file: src/abstract.tex
\begin{abstract}
This paper introduces an alternative approach to sampling from autoregressive models. Autoregressive models are typically sampled sequentially, according to the transition dynamics defined by the model. Instead, we propose a sampling procedure that initializes a sequence with white noise and follows a Markov chain defined by Langevin dynamics on the global log-likelihood of the sequence. This approach parallelizes the sampling process and generalizes to conditional sampling. Using an autoregressive model as a Bayesian prior, we can steer the output of a generative model using a conditional likelihood or constraints. We apply these techniques to autoregressive models in the visual and audio domains, with competitive results for audio source separation, super-resolution, and inpainting.
\end{abstract}

%% file: src/intro.tex
\section{Introduction}
\label{section:intro}

Neural autoregressive models \citep{larochelle2011neural} are a popular family of generative models, with wide-ranging applications in a variety of domains including audio \citep{oord2016wavenet,dhariwal2020jukebox}, images \citep{van2016pixel,salimans2017pixelcnn++,parmar2018image,razavi2019generating}, and text \citep{radford2019language,brown2020language}.
These models parameterize the conditional distribution over a token in an ordered sequence, given previous tokens in the sequence. The standard approach to sampling from an autoregressive model iteratively generates tokens, according to a conditional distribution over tokens defined by the model, conditioned on the partial sequence of previously generated tokens. We will refer to this approach to sampling as the ancestral sampler.

There are two major drawbacks to ancestral sampling that limit the usefulness of autoregressive models in practical settings. First, ancestral sampling has time complexity that scales linearly in the length of the generated sequence.
For data such as high-resolution images or audio, ancestral sampling from an autoregressive model (where the tokens are pixels or sound pressure readings respectively) can be impractically slow.
Second, ancestral sampling is frustratingly inflexible. It is easy to sample the second half of a sequence conditioned on the first, but filling in the first half a sequence conditioned on the second naively requires training a new model that reverses the ordering of tokens in the autoregressive factorization. Conditioning on arbitrary subsets of tokens for tasks such as inpainting or super-resolution seems beyond reach of autoregressive modeling.

This paper introduces an alternative, parallel and flexible (PnF) sampler for autoregressive models that can be parallelized and steered using conditioning information or constraints.\footnote{Code and examples of PnF sampling are available at:\\ \url{https://grail.cs.washington.edu/projects/pnf-sampling/}.}
Instead of sampling tokens sequentially, the PnF sampler initializes a complete sequence (with random tokens) and proceeds to increase the log-likelihood of this sequence by following a Markov chain defined by Langevin dynamics \citep{neal2011mcmc} on a smoothed log-likelihood. The smoothing temperature is cooled over time according to an annealing schedule informed by \citet{song2019generative,song2020improved}. Convergence time of this annealed Langevin dynamics is empirically independent of the sequence length and, generalizing \citet{jayaram2020source}, the PnF sampler can be applied to posterior log-likelihoods to incorporate conditional information into the sampling process.

\begin{figure*}[ht!]
\begin{center}
\vspace{-2mm}
\centerline{\includegraphics[width=.95\textwidth]{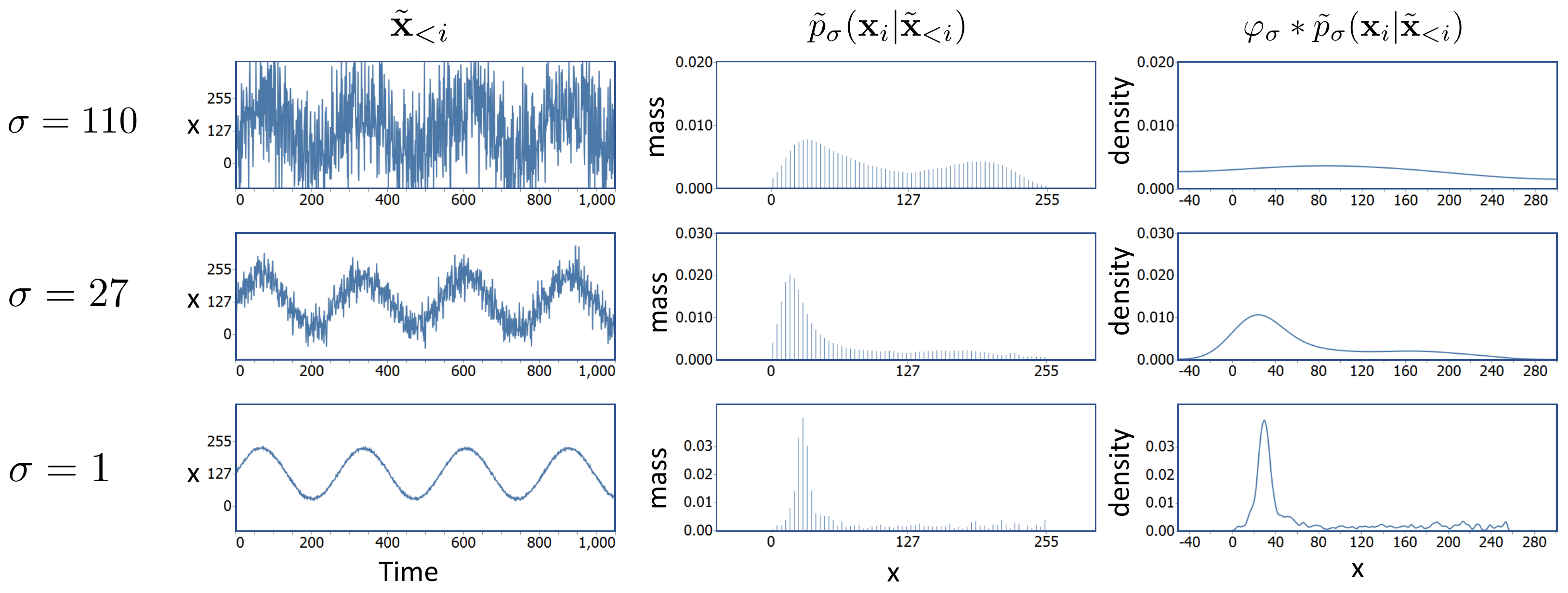}}
\vspace{-4mm}
\caption{A visual summary of discretized autoregressive smoothing. Given a noisy history $\tilde{\textbf{x}}_{<i} = \textbf{x}_{<i} + \bm\varepsilon_{<i}$ (left column) where $\bm\varepsilon \sim \mathcal{N}(0,\sigma^2I)$, we train a model to predict the un-noised distribution over $\textbf{x}_{i} \in \mathbb{R}$ (middle column). This distribution is discrete and non-differentiable in $\tilde{\textbf{x}}$; we convolve with a Gaussian $\varphi_\sigma(t) = \mathcal{N}(t;0,\sigma^2)$ to produce a continuous estimate of $\tilde{\textbf{x}}_{i}$ (right column). We can run Langevin dynamics on the continuous distribution, and gradually anneal the smoothing to approximate the target distribution.}
\label{figure:overview}
\vspace{-9mm}
\end{center}
\end{figure*}

The primary technical contribution of this paper is the development of the PnF sampler for discretized autoregressive models (Section \ref{sec:autoregressive}). Our interest in these models is motivated by their success as unconditional models of audio waves \citep{oord2016wavenet,mehri2017samplernn,dhariwal2020jukebox}.
Defined over a discrete lattice within a continuous space, these models occupy a middle ground between continuous and discrete models. For continuous models such as RNADE \citep{uria2013rnade}, PnF sampling can be directly applied as in \citet{song2019generative,jayaram2020source}. We defer the development of the PnF sampler for fully discrete models to future work.

In Section \ref{sec:sgld}, we present a stochastic variant of the PnF sampler based on stochastic gradient Langevin dynamics \citep{welling2011bayesian}. This is an embarrassingly parallel, asynchronous distributed algorithm for autoregressive sampling. Using a WaveNet model, we show in Section~\ref{sec:speed} that stochastic PnF sampling approximates the quality of ancestral sampling to arbitrary accuracy, with compute time that is inverse proportional to the number of computing devices. This allows PnF sampling to take full advantage of modern, massively parallel computing infrastructure.

We will see in Section \ref{sec:smoothing} how the PnF sampler can find solutions to general posterior sampling problems, using an unconditional generative model as a prior. In Section \ref{sec:experiments} we present applications of the PnF sampler to a variety of Bayesian image and audio inverse problems. We focus on linear inverse problems, using PixelCNN++ \citep{salimans2017pixelcnn++} and WaveNet \citep{oord2016wavenet} models as priors. Sections \ref{sec:source}, \ref{sec:super}, and \ref{sec:inpaint} demonstrate PnF conditional sampling for source separation, super-resolution, inpainting respectively.
PnF sampling results correlate strongly with the strength of the generative model used as a prior; as better autoregressive models are developed, they can be used with PnF sampling to improve performance on conditional generation tasks.
We refer the reader to the project website for demonstrations of audio PnF sampling.

%% file: src/related.tex
\section{Related Work}
\label{section:related}

The PnF autoregressive sampler is based on the annealed Langevin dynamics introduced in \citet{song2019generative}, which accelerates standard Langevin dynamics \citep{neal2011mcmc,du2019implicit} using a smoothing procedure in the spirit of simulated annealing \citep{kirkpatrick1983optimization} and graduated optimization \citep{blake1987visual}. The extension of annealed Langevin dynamics to conditional sampling problems was discussed in \citet{jayaram2020source} for source separation and image coloring problems, and developed further in \citet{song2020score} for general posterior sampling problems. The present work extends these methods to discretized autoregressive models, for which the smoothing procedures described in previous work are not directly applicable \citep{frank2020problems}. 
Markov-chain Monte Carlo posterior samplers based on Gibbs sampling rather than Langevin dynamics are proposed in \citet{theis2015generative} and \citet{hadjeres2017deepbach} as solutions for inpainting tasks.

The slow speed of ancestral sampling is a persistent obstacle to the adoption and deployment of autoregressive models. This has inspired algorithms that seek to parallelize the sampling process. Parallel WaveNet \citep{oord2018parallel} and ClariNet \citep{ping2019clarinet} train generative flow models to mimic the behavior of an autoregressive model. Sampling a flow model requires only one pass through a feed-forward network and can be distributed across multiple devices. \citet{wiggers2020predictive} and \citet{song2020nonlinear} propose fixed-point algorithms that, like PnF sampling, iteratively refine an initial sample from a simple distribution into a sample from the target distribution. But none of these methods are easily adaptable to source separation (Section \ref{sec:source}) or more general conditional sampling tasks.

Like anytime sampling \citep{xu2021anytime}, PnF sampling offers a tradeoff between sample quality and computational budget. The algorithm's iterates gradually mix to the target distribution and, by stopping early, we can approximate samples from this distribution using less computation. We explore the empirical tradeoff between sample quality and computation for PnF sampling from autoregressive models in Section~\ref{sec:quality}. The anytime sampler proposed in \citet{xu2021anytime} requires a specific model architecture based on the VQ-VAE \citep{van2017neural,razavi2019generating}. In contrast, the PnF sampler can be used with any likelihood-based model. Unlike an anytime sampler, the computational budget for PnF  sampling must be specified in advance: halting prior to completing the annealing schedule will result in noisy samples.

Bayesian inverse problems are explored extensively in theoretical settings, where the prior is given by a simple analytical distribution \citep{tropp2010computational,knapik2011bayesian,wang2017bayesian}. These problems have also been studied using learned priors given by GAN's, with a focus on linear inverse problems \citep{rick2017one,bora2017compressed,raj2019gan}. These GAN-based approaches are tailored to the latent variable architecture of the model, performing latent space optimizations to find codes that correspond to desired outputs. There is no obvious extension of these latent variable approaches to autoregressive models.

While we focus on autoregressive models, due to their strong empirical performance as unconditional models of audio, PnF sampling could be applied more generally with other likelihood-based models. In the audio space, this includes recent diffusion models \citep{kong2020diffwave, chen2020wavegrad}. Note however that audio vocoder models  \citep{prenger2019waveglow,kim2018flowavenet,ping2020waveflow}, which rely on spectrogram conditioning, cannot be adapted as priors for the source separation, super-resolution, and inpainting experiments presented in Section~\ref{sec:experiments}. In addition, GAN based models \citep{donahue2019adversarial, kumar2019melgan}, which are not likelihood based, cannot be sampled using PnF. 

%% file: src/method.tex
\section{Parallel and Flexible Sampling}
\label{sec:methods}

We want to sample from an autoregressive generative model over some indexed sequence of values $\textbf{x} \in \mathcal{X}^n$ where
\begin{equation}\label{eqn:autoregressive}
p(\textbf{x}) = \prod_{i=1}^n p(\textbf{x}_i|\textbf{x}_{<i}).
\end{equation}
We are particularly interested in developing a sampler for discretized autoregressive models, where $\mathcal{X}~=~\mathbb{R}$ and each conditional $p(\textbf{x}_i|\textbf{x}_{<i})$ has support on a finite set of scalar values $\mathcal{D} = \{e_1,\dots,e_d\} \subset \mathbb{R}$. The set $\mathcal{D}^n$ could represent, for example, an 8-bit encoding of an image or audio wave.

We propose to sample from $p(\textbf{x})$ via Langevin dynamics. Let $\textbf{x}_0~\sim~\text{Uniform}(\mathbb{R}^n)$, $\bm\varepsilon_t~\sim~\mathcal{N}(0,I_n)$, and define a Markov chain
\begin{align}\label{eqn:langevin}
&\textbf{x}^{(t+1)} \equiv \textbf{x}^{(t)} + \eta\nabla_\textbf{x} \log p(\textbf{x}^{(t)}) + \sqrt{2\eta}\bm\varepsilon_t
\end{align}
If $p(\textbf{x})$ were a smooth density then, for sufficiently small $\eta$, the Markov chain mixes and $\textbf{x}^{(t)}$ converges in distribution to $p$ as $t\to\infty$. But a discretized probability distribution defined over $\mathcal{D}^n$ is not smooth; the gradient $\nabla_\textbf{x} \log p(\textbf{x}^{(t)})$ is not even well-defined. In Section \ref{sec:autoregressive} we propose a smoothing of the discrete model $p(\textbf{x})$, creating a differentiable density on which the Markov chain \eqref{eqn:langevin} can mix.

To support conditional generation, we then turn our attention to sampling from the posterior of a joint distribution $p(\textbf{x},\textbf{y}) = p(\textbf{y}|\textbf{x})p(\textbf{x})$, where $p(\textbf{x})$ is an autoregressive model over $\textbf{x} \in \mathbb{R}^n$ and $p(\textbf{y}|\textbf{x})$ is a conditional likelihood. Langevin dynamics for sampling from the posterior $p(\textbf{x}|\textbf{y})$ are given by
\begin{align}\label{eqn:jointlangevin}
&\textbf{x}^{(t+1)} \equiv \textbf{x}^{(t)} + \eta\nabla_\textbf{x} \log p(\textbf{x}^{(t)}|\textbf{y}) + \sqrt{2\eta}\bm\varepsilon_t\\
&= \textbf{x}^{(t)} +\eta \nabla_\textbf{x} \left(\log p(\textbf{x}^{(t)}) + \log p(\textbf{y}|\textbf{x}^{(t)})\right)+ \sqrt{2\eta}\bm\varepsilon_t.\notag
\end{align}
This is a convenient Markov chain for posterior sampling, because the partition function $p(\textbf{y})$ vanishes when we take the gradient. However, like $p(\textbf{x})$, the posterior $p(\textbf{x}|\textbf{y})$ is not smooth. In Section \ref{sec:smoothing} we propose a smoothing of the joint distribution $p(\textbf{x},\textbf{y})$, for which the posterior $p(\textbf{x}|\textbf{y})$ is differentiable and the Markov chain \eqref{eqn:jointlangevin} can mix.

\begin{algorithm}[tb]
   \caption{Parallel and Flexible Sampling}
   \label{alg:fnf}
\begin{algorithmic}
   \STATE {\bfseries Input:} $\textbf{y}$, $\{\sigma_i\}_{i=1}^L$, $\delta$, $T$
   \STATE Sample $\textbf{x}^{(0)} \sim \mathcal{N}(0,\sigma_1^2I_n)$
   \FOR{$i \gets 1$ {\bfseries to} $L$}
   \STATE $\eta_i \gets \delta \cdot \sigma_i^2/\sigma_L^2$
   \FOR{$t=1$ {\bfseries to} $T$}
       \STATE Sample $\bm\varepsilon_t \sim \mathcal{N}(0,I_n)$
       \STATE $\textbf{g}^{(t)} \gets \nabla_\textbf{x} \log p_{\sigma_i}(\textbf{x}^{(t)}) + \nabla_\textbf{x} \log p_{\sigma_i}(\textbf{y}|\textbf{x}^{(t)})$
       \STATE $\textbf{x}^{(t+1)} \gets \textbf{x}^{(t)}+ \eta_i \textbf{g}^{(t)} + \sqrt{2\eta}\bm\varepsilon_t$
   \ENDFOR
   \ENDFOR
\end{algorithmic}
\end{algorithm}
\setlength{\textfloatsep}{0.3cm}
\setlength{\floatsep}{0.3cm}

Given a smoothing procedure parameterized by a temperature parameter, we appeal to the simulated annealing heuristic developed in \citet{song2019generative} to turn down the temperature as the Markov chain \eqref{eqn:langevin} or \eqref{eqn:jointlangevin} mixes. In contrast to classical Markov chain sampling, for which samples $\textbf{x}^{(t)}$ converge in distribution to $p$, annealed Langevin dynamics converges asymptotically to a single point distributed approximately according to $p$. Algorithm~\ref{alg:fnf} describes these annealed Langevin dynamics given a smoothed prior $p_\sigma(\textbf{x})$ and smoothed likelihood $p_\sigma(\textbf{y}|\textbf{x})$. The structure of this algorithm is the same as the annealed Langevin dynamics presented in \citet{song2019generative} (the unconditional case where $\nabla_\textbf{x} p_\sigma(\textbf{y}|\textbf{x}) = 0$) and \citet{jayaram2020source} (the conditional case); the novel contribution of this paper is the smoothing algorithm for evaluating $\nabla_\textbf{x} p_\sigma(\textbf{x})$ given $p(\textbf{x})$ (Section~\ref{sec:autoregressive}) and $\nabla_\textbf{x} p_\sigma(\textbf{y}|\textbf{x})$ given $p(\textbf{y}|\textbf{x})$ (Section~\ref{sec:smoothing}).

Each step of the Langevin dynamics described in Equations \eqref{eqn:langevin} or \eqref{eqn:jointlangevin} requires inference of $\log p(\textbf{x})$, an $O(n)$ operation. Unlike sequential sampling, calculating $\log p(\textbf{x})$ for a given sequence $\textbf{x} \in \mathbb{R}^n$ is embarrassingly parallel; for moderate sequence lengths $n$, the cost of computing $\log p(\textbf{x})$ is essentially constant using a modern parallel computing device. When $n$ is very large, e.g. for WaveNet models where just a minute of audio has $n > 10^6$ samples, it can be convenient to distribute sampling across multiple computing devices. In Section \ref{sec:sgld} we describe a stochastic variant of PnF sampling based on stochastic gradient Langevin dynamics that is easily distributed across a cluster of devices.

\subsection{Autoregressive Smoothing}\label{sec:autoregressive}

We consider models that parameterize the conditional distribution $p(\textbf{x}_i|\textbf{x}_{<i})$ with a categorical softmax distribution over $d = |\mathcal{D}|$ discrete values; given functions $f_i : \mathcal{X}^i \to \mathbb{R}^d$, we define
\begin{equation}\label{eqn:softmax}
p(\textbf{x}_i = e_k|\textbf{x}_{<i}) = \frac{\exp(f_{i,k}(\textbf{x}_{<i}))}{\sum_{\ell=1}^d \exp(f_{i,\ell}(\textbf{x}_{<i}))}.
\end{equation}
The functions $f_i$ are typically given by a neural network, with shared weights across the sequential indices $i$. Collectively, these conditional models define the joint distribution $p(\textbf{x})$ according to Equation \eqref{eqn:autoregressive}.

We cannot directly compute gradients of the distribution $p(\textbf{x})$ defined by discrete conditionals $p(\textbf{x}_i|\textbf{x}_{<i})$. Instead, we smooth $p(\textbf{x})$ by convolving it with a spherical Gaussian $\mathcal{N}(0,\sigma^2 I_n)$. This smoothing relies on the fact that $e_k \in \mathcal{D}$ represent scalar values on the real line, and therefore the discrete distribution $p(\textbf{x}_i|\textbf{x}_{<i})$ can be viewed as a linear combination of weighted Dirac spikes on $\mathcal{D} \subset \mathbb{R}$. If $\phi_\sigma(\textbf{x})$ denotes the density of a spherical Gaussian $\mathcal{N}(0,\sigma^2 I_n)$ on $\mathbb{R}^n$ then we define a density $p_\sigma(\tilde{\textbf{x}})$ on $\mathbb{R}^n$ given by
\begin{equation}\label{eqn:smoothprior}
p_\sigma(\tilde{\textbf{x}}) = (\phi_\sigma * p)(\tilde{\textbf{x}}) = \int \phi_\sigma(\tilde{\textbf{x}} - \textbf{x})p(\textbf{x})\,d\textbf{x}.
\end{equation}
This distribution has well-defined gradients $\nabla_{\tilde{\textbf{x}}}\log p_\sigma(\tilde{\textbf{x}})$ and $p_\sigma(\tilde{\textbf{x}}) \to p(\textbf{x})$ in total variation as $\sigma^2 \to 0$.

When $p(\textbf{x})$ is a deep autoregressive model, the convolution \eqref{eqn:smoothprior} is difficult to calculate directly. Previous work proposed training a smoothed model $p_\sigma(\tilde{\textbf{x}})$ by fine-tuning a model $p(\textbf{x})$ on noisy data $\tilde{\textbf{x}} = \textbf{x} + \bm\varepsilon_\sigma$ where $\textbf{x} \sim p$ and $\bm\varepsilon_\sigma \sim \mathcal{N}(0,\sigma^2I_n)$ \citep{jayaram2020source}. This approach cannot be directly applied to discrete autoregressive models as defined by Equation \eqref{eqn:softmax}. The obstruction is that noisy samples $\tilde{\textbf{x}}_i \in \mathbb{R}$ are not supported by the discretization $\mathcal{D}$.
One way to address the problem is to replace the discrete model $p(\textbf{x})$ with a continuous autoregressive model of $p_\sigma(\tilde{\textbf{x}})$, e.g. RNADE \citep{uria2013rnade}. We avoid this approach because fine-tuning $p(\textbf{x})$ to $p_\sigma(\tilde{\textbf{x}})$ becomes complicated when these models have different architectures.

Instead of directly fine-tuning $p(\textbf{x})$ to a model $p_\sigma(\tilde{\textbf{x}})$, we combine an analytic calculation with an auxiliary model learned via fine-tuning. Let $\tilde{p}_\sigma(\textbf{x}_i|\tilde{\textbf{x}}_{<i})$ denote a (discrete) conditional model trained to predict $\textbf{x}_i$ given noisy covariates $\tilde{\textbf{x}}_{<i} = \textbf{x}_{<i} + \bm\varepsilon_{\sigma,<i}$. If $\varphi_\sigma$ denotes the density of $\mathcal{N}(0,\sigma^2)$ then we can re-write the factored density $p_\sigma(\tilde{\textbf{x}})$ as
\begin{equation}
p_\sigma(\tilde{\textbf{x}}) = \prod_{i=1}^n p_\sigma(\tilde{\textbf{x}}_i|\tilde{\textbf{x}}_{<i}) = \prod_{i=1}^n \big(\varphi_\sigma*\tilde{p}_\sigma({}\cdot{}|\tilde{\textbf{x}}_{<i})\big)(\tilde{\textbf{x}}_i).
\end{equation}
On the right-hand side, we decompose the smoothed conditional densities $p_\sigma(\tilde{\textbf{x}}_i|\tilde{\textbf{x}}_{<i})$ into Gaussian convolutions of discrete conditionals $\tilde{p}_\sigma({}\cdot{}|\tilde{\textbf{x}}_{<i})$ evaluated at $\tilde{\textbf{x}}_i$. This suggests the following approach to evaluating $p_\sigma(\tilde{\textbf{x}}_i|\tilde{\textbf{x}}_{<i})$:
\begin{itemize}
\item Learn a (discrete) model $\tilde{p}_\sigma(\textbf{x}_i|\tilde{\textbf{x}}_{<i})$, trained to predict the un-noised value $\textbf{x}_i$ given noisy history $\tilde{\textbf{x}}_{<i}$. This model can be learned efficiently by finetuning a pre-trained model $p(\textbf{x}_i|\textbf{x}_{<i})$ on noisy covariates $\tilde{\textbf{x}}_{<i}$. This is visualized in the middle column of Figure~\ref{figure:overview}.

\item Evaluate the Gaussian convolution $\varphi_\sigma*\tilde{p}_\sigma({}\cdot{}|\tilde{\textbf{x}}_{<i})$ at $\tilde{\textbf{x}}_i$ to compute $p_\sigma(\tilde{\textbf{x}}_i|\tilde{\textbf{x}}_{<i})$. This convolution can be calculated in closed form given $\tilde{p}_\sigma(\textbf{x}_i|\tilde{\textbf{x}}_{<i})$. This is visualized in the right column of Figure~\ref{figure:overview}.
\end{itemize}

The convolution $p_\sigma(\tilde{\textbf{x}}_i|\tilde{\textbf{x}}_{<i}) = \big(\varphi_\sigma*\tilde{p}_\sigma({}\cdot{}|\tilde{\textbf{x}}_{<i})\big)(\tilde{\textbf{x}}_i)$ has a simple closed form given by a Gaussian mixture model
\begin{equation}
p_\sigma(\tilde{\textbf{x}}_i|\tilde{\textbf{x}}_{<i}) = \sum_{k=1}^d \tilde{p}_\sigma(e_k|\tilde{\textbf{x}}_{<i}) \varphi_\sigma(\tilde{\textbf{x}}_i - e_k).
\end{equation}

Using the softmax parameterization of $\tilde{p}(\textbf{x}_i|\textbf{x}_{<i})$ given by Equation \eqref{eqn:softmax}, with fine-tuned logits $f_{\sigma,i} : \mathcal{X}^{\otimes i} \to \mathbb{R}^d$, the log-density of this smoothed conditional density can be written in a numerically stable form:
\begin{align}
&\log\hspace{1mm} p_\sigma(\textbf{x}_i|\textbf{x}_{<i}) = - \log \sum_{\ell=1}^d \exp(f_{\sigma,i,\ell}(\textbf{x}_{<i}))\\
&+ \log \sum_{k=1}^d \exp\left(f_{\sigma,i,k}(\textbf{x}_{<i})-\frac{1}{2\sigma^2}(\textbf{x}_i - e_k)^2\right) + C.\notag
\end{align}

This smoothing procedure requires us to store finetuned models $\tilde{p}_\sigma(\textbf{x}_i|\tilde{\textbf{x}}_{<i})$ for each of the $L$ noise levels $\sigma \in \{\sigma_1,\dots,\sigma_L\}$. This could be avoided by training a single noise-conditioned generative model. The advantage of finetuning is that we can directly use standard models, without any adjustment to the network architecture or subsequent hyper-parameter tuning of the modified architecture; this cleanly decouples our approach to conditional sampling from neural architecture design questions. Note that while we store $L$ copies of the model, there is no additional memory overhead: these models are loaded and unloaded serially during optimization as we anneal the noise levels, so only one model is resident in memory at a time. While memory is a scarce resource, disk space is generally abundant.

\subsection{Stochastic Gradient Langevin Dynamics}\label{sec:sgld}

\begin{algorithm}[tb]
   \caption{Stochastic Parallel and Flexible Sampling}
   \label{alg:sfnf}
\begin{algorithmic}
   \STATE {\bfseries Input:} $\textbf{y}$, $\{\sigma_i\}_{i=1}^L$, $\delta$, $T$
   \STATE Sample $\textbf{x} \sim \mathcal{N}(0,\sigma_1^2I_n)$
   \FOR{$i \gets 1$ {\bfseries to} $L$}
   \STATE $\eta_i \gets \delta \cdot \sigma_i^2/\sigma_L^2$
   \STATE Fork()
   \FOR{$t=1$ {\bfseries to} $T$}
       \STATE Sample $j \sim \text{Uniform}\{1,\dots,n\}$
       \STATE Sample $\bm\varepsilon_t \sim \mathcal{N}(0,I_c)$
       \STATE Read $\textbf{x}^{(t)}_{j-w:j+c+w} \gets \textbf{x}_{j-w:j+c+w}$
       \STATE $\textbf{g}_j^{(t)} \gets \nabla_{\textbf{x}_{j:j+c}} \log p_{\sigma_i}(\textbf{x}_{j-w:j+c+w}^{(t)})$
       \STATE \hspace{7mm}$+\hspace{1mm} \nabla_{\textbf{x}_{j:j+c}} \log p_{\sigma_i}(\textbf{y}_{j:j+c}^{(t)}|\textbf{x}_{j-w:j+c+w}^{(t)})$
       \STATE Write $\textbf{x}_{j:j+c} \gets \textbf{x}_{j:j+c}^{(t)}+ \eta_i \textbf{g}_j^{(t)} + \sqrt{2\eta}\bm\varepsilon_t$
   \ENDFOR
   \STATE Synchronize()
   \ENDFOR
\end{algorithmic}
\end{algorithm}
\setlength{\textfloatsep}{0.3cm}
\setlength{\floatsep}{0.3cm}

Calculating the Langevin updates described in Equations \eqref{eqn:langevin} and \eqref{eqn:jointlangevin} requires $O(n)$ operations to compute $\log p_\sigma(\textbf{x})$, given a sequence $\textbf{x} \in \mathbb{R}^n$. This calculation decomposes into $n$ calculations of $\log p_\sigma(\textbf{x}_i|\textbf{x}_{<i})$ ($i = 1,\dots,n$) which can each be computed in parallel (this is what allows autoregressive models to be efficiently trained using the maximum likelihood objective). But for very large values of $n$, even modern parallel computing devices cannot fully parallelize all $n$ calculations. In this section, we develop a stochastic variant of PnF sampling (Algorithm \ref{alg:sfnf}) that is easily distributed across multiple devices.

Instead of making batch updates on a full sequence $\textbf{x} \in \mathbb{R}^n$, consider updating a single coordinate $j \in \{1,\dots,n\}$:
\begin{equation}
\textbf{x}^{(t+1)}_j = \textbf{x}^{(t)}_j + \eta \nabla_{\textbf{x}_j} \log p_\sigma(\textbf{x}^{(t)}) + \sqrt{2\eta}\bm\varepsilon^{(t)}_j.
\end{equation}
This coordinate-wise derivative is only dependent on the tail of the sequence $\textbf{x}_{\geq j}$:
\begin{equation}
\nabla_{\textbf{x}_j} \log p_\sigma(\textbf{x}) = \sum_{i=j}^n \nabla_{\textbf{x}_j} \log p_\sigma(\textbf{x}_i|\textbf{x}_{<i}).
\end{equation}
This doesn't look promising; calculating an update on a single coordinate $\textbf{x}_j$ required $n-j$ inference calculations $p_\sigma(\textbf{x}_i|\textbf{x}_{<i})$. But models over very long sequences, including WaveNets, usually make a Markov assumption $p(\textbf{x}_i|\textbf{x}_{<i}) = p(\textbf{x}_i|\textbf{x}_{i-w},\dots,\textbf{x}_{i-1})$ for some limited contextual window of length $w$. In this case, the coordinate-wise derivative requires only $w$ calls:
\begin{equation}
\nabla_{\textbf{x}_j} \log p_\sigma(\textbf{x}) = \sum_{i=j}^{j+w} \nabla_{\textbf{x}_j} \log p_\sigma(\textbf{x}_i|\textbf{x}_{<i}).
\end{equation}

Calculating a gradient on a contiguous block of $c$ coordinates leads to a more efficient update
\begin{equation}\label{eqn:block}
\nabla_{\textbf{x}_{j:j+c}} \log p_\sigma(\textbf{x}) = \sum_{i=j}^{j+c+w} \nabla_{\textbf{x}_{j:j+c}} \log p_\sigma(\textbf{x}_i|\textbf{x}_{<i}).
\end{equation}
Calculating Equation \eqref{eqn:block} requires transmission of a block $\{\textbf{x}_{j-w},\dots,\textbf{x}_{j+c+w}\}$ of length $c + 2w$ to the computing device, and $c+w$ calculations $p_\sigma(\textbf{x}_i|\textbf{x}_{<i})$ in order to compute the gradient of a block of length $c$. If we partition a sequence of length $n$ into $n/c$ blocks of length $c$, then we can distribute computation of $\nabla_{\textbf{x}} \log p_\sigma(\textbf{x})$ with an overhead factor of $1 + w/c$. This motivates choosing $c$ as large as possible, under the constraint that $c+w$ calculations $p_\sigma(\textbf{x}_i|\textbf{x}_{<i})$ can still be parallelized on a single device.

We can calculate $\nabla_{\textbf{x}} \log p_\sigma(\textbf{x})$ by aggregating $n/c$ blocks of gradients according to Equation \eqref{eqn:block}, requiring synchronous communication between $n/c$ machines for every update Equation \eqref{eqn:langevin}; this is a MapReduce algorithm \citep{dean2004mapreduce}. We propose a bolder approach in Algorithm~\ref{alg:sfnf} based on block-stochastic Langevin dynamics \citep{welling2011bayesian}. If $j \in \{1,\dots,n\}$ is chosen uniformly at random then Equation \eqref{eqn:block} is an unbiased estimate of $\nabla_{\textbf{x}} \log p_\sigma(\textbf{x})$. This motivates block-stochastic updates on patches, which multiple devices can perform asynchronously, a Langevin analog to Hogwild! \citep{niu2011hogwild}.

\subsection{Smoothing a Joint Distribution}\label{sec:smoothing}

We now consider joint distributions $p(\textbf{x},\textbf{y}) = p(\textbf{y}|\textbf{x})p(\textbf{x})$ over sources $\textbf{x} \in \mathcal{X}^n$ and measurements $\textbf{y} \in \mathcal{Y}^n$. For example, $\textbf{y}$ could be a low resolution version of the signal, or an observed mixture of two signals. We are particularly interested in measurement models of the form $\textbf{y} = g(\textbf{x})$, for some linear function $g : \mathcal{X}^n \to \mathcal{Y}^n$. We can view these measurements $\textbf{y}$ as degenerate likelihoods $p(\textbf{y}|\textbf{x})$ of the form
\begin{equation}\label{eqn:dirac}
p(\textbf{y}|\textbf{x}) = \delta(\textbf{y} - g(\textbf{x})),
\end{equation}
where $\delta$ denotes the Dirac delta function. This family of linear measurement models describes the source separation, in-painting, and super-resolution tasks featured in Section~\ref{sec:experiments}, as well as other linear inverse problems including sparse recovery and image colorization. 

Extending our analysis in Section \ref{sec:autoregressive}, we can smooth the joint density $p(\textbf{x},\textbf{y})$ by convolving $\textbf{x}$ with a spherical Gaussian $\mathcal{N}(0,\sigma^2 I_n)$. Let $\tilde{\textbf{x}} = \textbf{x} + \bm\varepsilon_\sigma$ where $\bm\varepsilon_\sigma \sim \mathcal{N}(0,\sigma^2 I_n)$. Note that $\tilde{\textbf{x}}$ is conditionally independent of $\textbf{y}$ given $\textbf{x}$ and therefore the joint distribution over $\textbf{x}$, $\tilde{\textbf{x}}$, and $\textbf{y}$ can be factored as
\begin{equation}\label{eqn:smoothdensity}
p_\sigma(\textbf{x},\textbf{y},\tilde{\textbf{x}}) = p_\sigma(\tilde{\textbf{x}}|\textbf{x})p(\textbf{y}|\textbf{x})p(\textbf{x}).
\end{equation}
We will work with the smoothed marginal $p_\sigma(\tilde{\textbf{x}},\textbf{y})$ of the joint distribution $p_\sigma(\textbf{x},\textbf{y},\tilde{\textbf{x}})$. If $\phi_\sigma(\textbf{x})$ denotes the density of a spherical Gaussian $\mathcal{N}(0,\sigma^2 I_n)$ on $\mathbb{R}^n$ then the marginal $p_\sigma(\tilde{\textbf{x}},\textbf{y})$ of Equation \eqref{eqn:smoothdensity} can be expressed by
\begin{equation}\label{eqn:jointconv}
p_\sigma(\tilde{\textbf{x}},\textbf{y}) = \int \phi_\sigma(\tilde{\textbf{x}} - \textbf{x}) p(\textbf{x},\textbf{y})\,d\textbf{x}
\end{equation}
This density approximates the original distribution $p(\textbf{x},\textbf{y})$ in the sense that $p_\sigma(\tilde{\textbf{x}},\textbf{y}) \to p(\textbf{x},\textbf{y})$ in total variation as $\sigma^2 \to 0$.

The smoothed density $p_\sigma(\tilde{\textbf{x}},\textbf{y})$ can be factored as $p_\sigma(\textbf{y}|\tilde{\textbf{x}})p_\sigma(\tilde{\textbf{x}})$. The density $p_\sigma(\tilde{\textbf{x}})$ is simply the smoothed Gaussian convolution of $p(\textbf{x})$ given by Equation \eqref{eqn:smoothprior}.
For general likelihoods $p(\textbf{y}|\textbf{x})$, because $\textbf{y}$ is conditionally independent of $\tilde{\textbf{x}}$ given $\textbf{x}$, we can write the density $p_\sigma(\textbf{y}|\tilde{\textbf{x}})$ by marginalizing over $\textbf{x}$ as
\begin{align}
p_\sigma(\textbf{y}|\tilde{\textbf{x}}) &= \int p_\sigma(\textbf{y}|\tilde{\textbf{x}},\textbf{x})p_\sigma(\textbf{x}|\tilde{\textbf{x}})\,d\textbf{x}\\
&= \int p(\textbf{y}|\textbf{x})\phi_\sigma(\textbf{x} - \tilde{\textbf{x}})\,d\textbf{x}.\label{eqn:generalsmoothing}
\end{align}
This integral is difficult to calculate directly. One way to evaluate $p_\sigma(\textbf{y}|\tilde{\textbf{x}})$ is to take the same approach described in Section \ref{sec:autoregressive} to evaluate $p_\sigma(\tilde{\textbf{x}})$: fine-tune the model $p(\textbf{y}|\textbf{x})$ to a model $\tilde p(\textbf{y}|\tilde{\textbf{x}})$ that predicts $\textbf{y}$ given noisy covariates $\tilde{\textbf{x}} = \textbf{x} + \bm\varepsilon_\sigma$. A similar procedure (using a noise-conditioned architecture rather than finetuning) is described in Section 5 of \citet{song2020score}.

For linear measurement models with the form given by Equation \eqref{eqn:dirac}, the smoothed density $p_\sigma(\textbf{y}|\tilde{\textbf{x}})$ can be calculated in closed form. Writing the linear function $g : \mathcal{X}^n \to \mathcal{Y}^n$ as a matrix $g(\textbf{x}) = A\textbf{x}$, we have
\begin{equation}\label{eqn:smoothlikelihood}
\textbf{y} = g(\textbf{x}) = g(\tilde{\textbf{x}}) + g(-\bm\varepsilon_\sigma) \sim \mathcal{N}\left(g(\tilde{\textbf{x}}),\sigma^2AA^T\right).
\end{equation}

This smoothing given by Equation \eqref{eqn:smoothlikelihood} generalizes the smoothing proposed in \citet{jayaram2020source} for source separation. That work proposed separately smoothing the prior and likelihood, resulting in a smoothed likelihood $p(\tilde{\textbf{y}}|\textbf{x})$ over $\tilde{\textbf{y}} = \textbf{y} + \bm\varepsilon_\textbf{y}$, where $\bm\varepsilon_\textbf{y} \sim \mathcal{N}(0,\sigma^2 I_n)$. This is equivalent to Equation \eqref{eqn:smoothlikelihood} in the case of source separation, for which $g(\textbf{x}) = \frac{1}{2}\textbf{x}_1 + \frac{1}{2}\textbf{x}_2$ and therefore $g(\textbf{x}) \sim \mathcal{N}\left(g(\tilde{\textbf{x}}),\sigma^2I\right)$.

For general likelihoods $p(\textbf{y}|\textbf{x})$ (e.g. a classifier) the conditioning values $\textbf{y}$ may depend on the whole sequence $\textbf{x}$. In this case, stochastic PnF must read the entire sequence $\textbf{x}$ in order to calculate the posterior
\begin{equation}
\hspace{-1mm}\nabla_{\textbf{x}_{j:j+c}} \log p(\textbf{x}|\textbf{y}) = \nabla_{\textbf{x}_{j:j+c}} \big(\log p(\textbf{x}) + \log p(\textbf{y}|\textbf{x})\big).
\end{equation}
But for long sequences $\textbf{x}$ such as audio, the conditioning information $\textbf{y}$ is often a local function of the sequence $\textbf{x}$. In this case, $\textbf{x} \in \mathcal{X}^n$, $\textbf{y} \in \mathcal{Y}^n$, $\textbf{y}_i = g(\textbf{x}_{N(i)})$ where $N(i)$ is a local neighborhood of indices near $i$, and the likelihood decomposes via conditional independence into
\begin{equation}
\log p(\textbf{y}|\textbf{x}) = \sum_{i=1}^n \log p(\textbf{y}_i|\textbf{x}_{N(i)}).
\end{equation}
All experiments presented in Section \ref{sec:experiments} feature this conditioning pattern. For spectrogram conditioning, $N(i)$ is the set of indices (centered at $i$) required to compute a short-time Fourier transform. For source separation, super-resolution, and in-painting, $N(i) = i$. This allows us to compute block gradients of the conditional likelihood (Algorithm \ref{alg:sfnf}).

%% file: src/experiments.tex
\section{Experiments}
\label{sec:experiments}

We present qualitative and quantitive results of PnF sampling for WaveNet models of audio \citep{oord2016wavenet} and a PixelCNN++ model of images \citep{salimans2017pixelcnn++}.
In Section \ref{sec:quality} we show that PnF sampling can produce samples of comparable quality to ancestral sampling. In Section \ref{sec:speed} we show that stochastic PnF sampling is faster than ancestral sampling, when parallelized across a modest number of devices. We go on to demonstrate how PnF sampling can be applied to a variety of image and audio restoration tasks: source separation (Section \ref{sec:source}), super-resolution (Section \ref{sec:super}), and inpainting (Section \ref{sec:inpaint}). We encourage the reader to browse the supplementary material  for qualitative examples of PnF audio sampling.

\subsection{Datasets}
For audio experiments we use the VCTK dataset \citep{vctk} consisting of 44 hours of speech, as well as the Supra Piano dataset \citep{shi2019supra} consisting of 52 hours of piano recordings. We use a random 80-20 train-test split of VCTK speakers and piano recordings for evaluation. Audio sequences are sampled at a $22$kHz, with $8$-bit $\mu$-law encoding \citep{recommendation1988pulse}, except for source separation where $8$-bit linear encoding is used. Sequences used for quantitative evaluation are $50$k sample excerpts, approximately 2.3 seconds of audio, chosen randomly from the longer test set recordings. For image experiments we use the CIFAR-10 dataset \citep{krizhevsky09learning} with the standard train-test split. Additional training and hyperparameter details can be found in the appendix.

\subsection{Quality of Generated Samples}\label{sec:quality}

\begin{figure*}[t]
\begin{minipage}[c]{.49\textwidth}
\begin{center}
\includegraphics[width=\columnwidth]{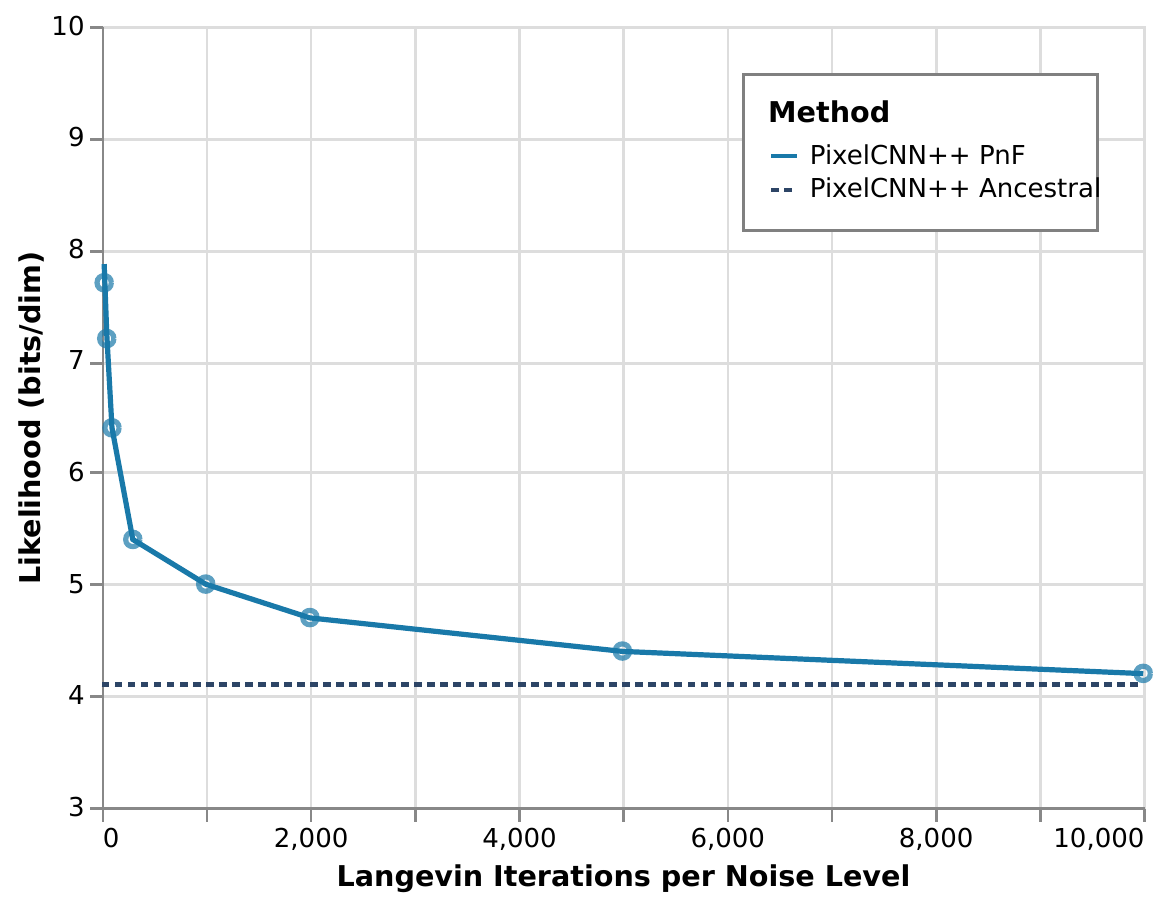}
\end{center}
\end{minipage}
\begin{minipage}[c]{.49\textwidth}
\begin{center}
\includegraphics[width=\columnwidth]{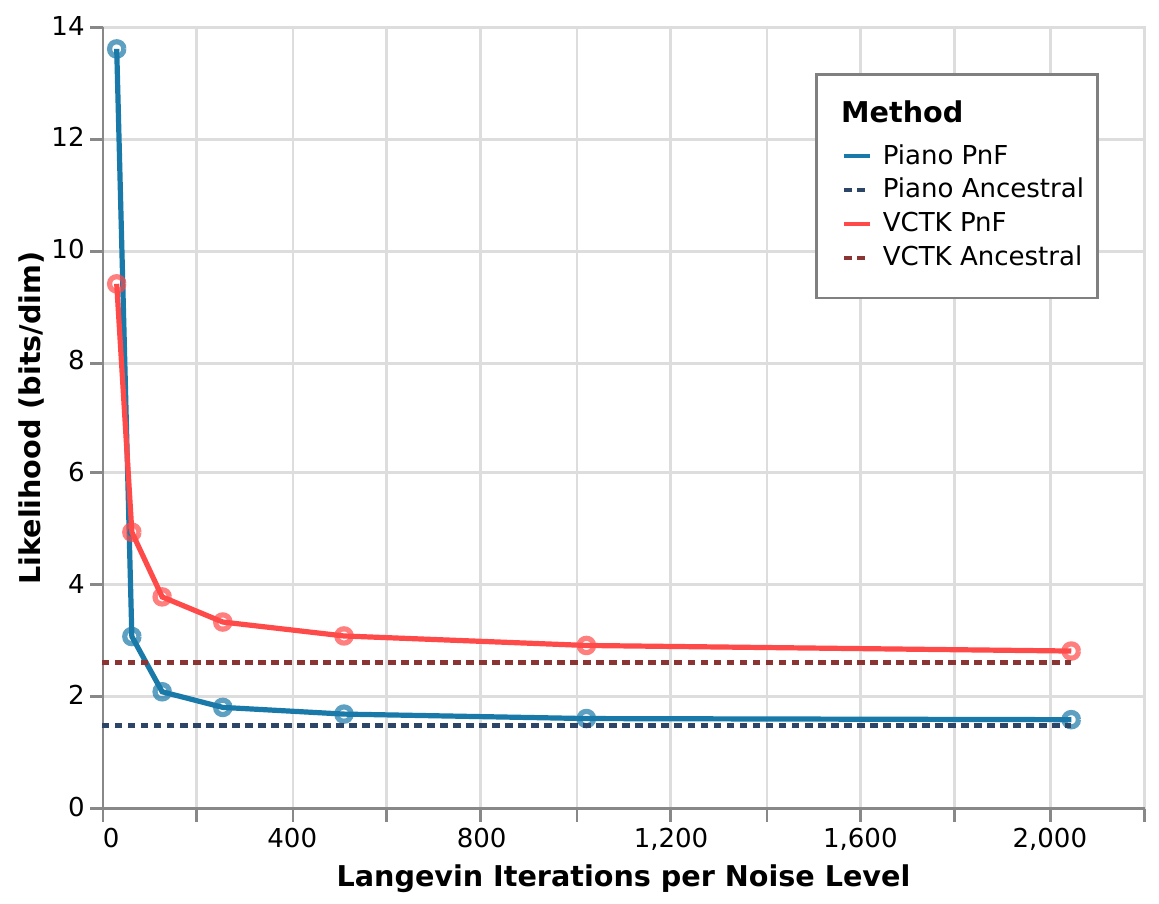}
\end{center}
\end{minipage}
\caption{As the number of Langevin iterations $T$ increases, the log-likelihood of sequences generated by PnF sampling approaches the log-likelihood of test set sequences. Left: sampling from a PixelCNN++ model trained on CIFAR-10. Right: sampling from WaveNet models trained on the Supra Piano and VCTK speech datasets. }
\label{fig:likelihood}
\vspace{-2mm}
\end{figure*}

To evaluate the quality of samples generated by PnF sampling, we follow a similar procedure to \citet{holtzman2020curious}. We compare log-likelihoods, calculated using the noiseless model $p(\textbf{x})$, of sequences generated by PnF sampling to sequences generated by ancestral sampling from the lowest-noise model $p_{\sigma_L}(\textbf{x})$. Because PnF-sampled sequences are continuous, we quantize these samples to 8-bit values when evaluating their likelihood under the noiseless model.
We consider PnF sampling to be successful if it generates sequences with comparable log-likelihoods to ancestral generations.

In Figure~\ref{fig:likelihood} we present quantitative results for PnF sampling using an unconditional PixelCNN++ model of CIFAR-10, and a spectrogram-conditioned WaveNet model of both voice and piano datasets. We evaluate $1,000$ generations (length $n = 50,000$ sequences for the WaveNet models) using various numbers of Langevin iterations, and report the median log likelihood of quantizations of these sequences under the noiseless model. Asymptotically, as the iterations $T$ of Langevin dynamics increase, the likelihood of PnF samples approaches the likelihood of ancestral samples. Audio PnF samples for various $T$ are presented on the project website.

\subsection{Speed and Parallelism}\label{sec:speed}

\begin{figure}[h]
\vspace{-1.5mm}
\begin{center}
\centerline{\includegraphics[width=\columnwidth]{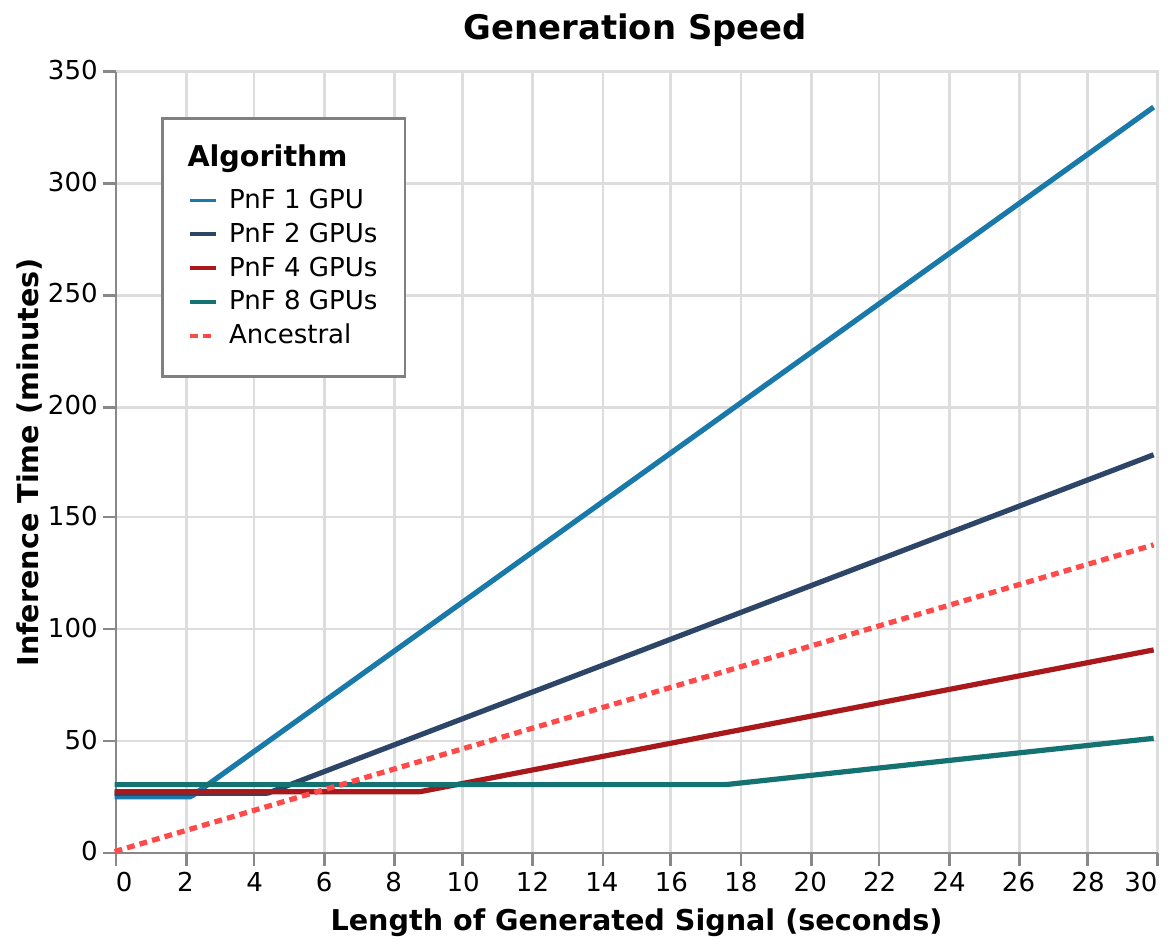}}
\caption{PnF sampling can be parallelized across multiple devices, resulting in faster inference time than ancestral sampling. Beyond a threshold level of computation, PnF sampling time is inversely proportional to the number of devices.}
\label{fig:speed}
\vspace{-4mm}
\end{center}
\end{figure}

Ancestral sampling has $O(n)$ serial runtime in the length $n$ of the generated sequence. Using the stochastic PnF sampler described in Section \ref{sec:sgld}, the serial runtime is $O(T)$, where $T$ is the number of Langevin iterations at each level of smoothing. We find empirically that we can set $T$ independent of $n$, so in principle the serial runtime of stochastic PnF is constant as a function of sequence length. In practice, we do not have an infinite supply of parallel devices, so the serial runtime of stochastic PnF grows inversely proportional to the number of devices. This behavior is demonstrated in Figure~\ref{fig:speed} for spectrogram-conditioned WaveNet stochastic PnF sampling using a cluster of $8$ Nvidia Titan Xp GPU's and $T=256$.
Each GPU can calculate Equation~\eqref{eqn:block} for a block of $c = 50,000$ samples (2.3 seconds of audio).
For PixelCNN++, we find that $T > n$ and therefore the PnF sampler does not improve sampling speed for this model.

Stochastic PnF sampling depends upon asynchronous writes being sparse so that memory overwrites, when two workers update overlapping blocks, are rare. This situation is analogous to the sparse update condition required for Hogwild! If blocks are length $c$ and the number of devices is substantially less than $n/c$, then updates are sufficiently sparse. But if the number of devices is larger than $n/c$, memory overwrites become common, and stochastic PnF sampling fails to converge. This imposes a floor on generation time determined by $c$, exhibited in Figure \ref{fig:speed}. We cannot substantially reduce this floor by decreasing $c$ because of the tradeoff between $c$ and the model's Markov window $w$ described in Section~\ref{sec:sgld}.

In general, PnF sampling becomes faster than ancestral sampling for long sequences $n$, in which case the stochastic variant of PnF sampling becomes necessary in order to distribute the calculation of $n$ conditional likelihoods. For shorter sequences, accurate unconditional samples can be produced more quickly with the ancestral sampler. Unconditional $32\times 32$ CIFAR-10 generation using the PixelCNN++ mode requires $T = 10,000$ Langevin iterations per noise level (Figure \ref{fig:likelihood}) for accurate samples; annealing through $L = 20$ levels requires a total of $L\times T = 200,000$ serial queries to the PixelCNN++ model, far more than $n = 3\times 32\times 32$ serial queries to PixelCNN++ for ancestral sampling. Note also that PixelCNN++ conditions on a full image ($w = n$) so the stochastic variant of Pnf sampling is not applicable to this model.

\begin{figure*}[ht!]
\begin{center}
\centerline{\includegraphics[width=\textwidth]{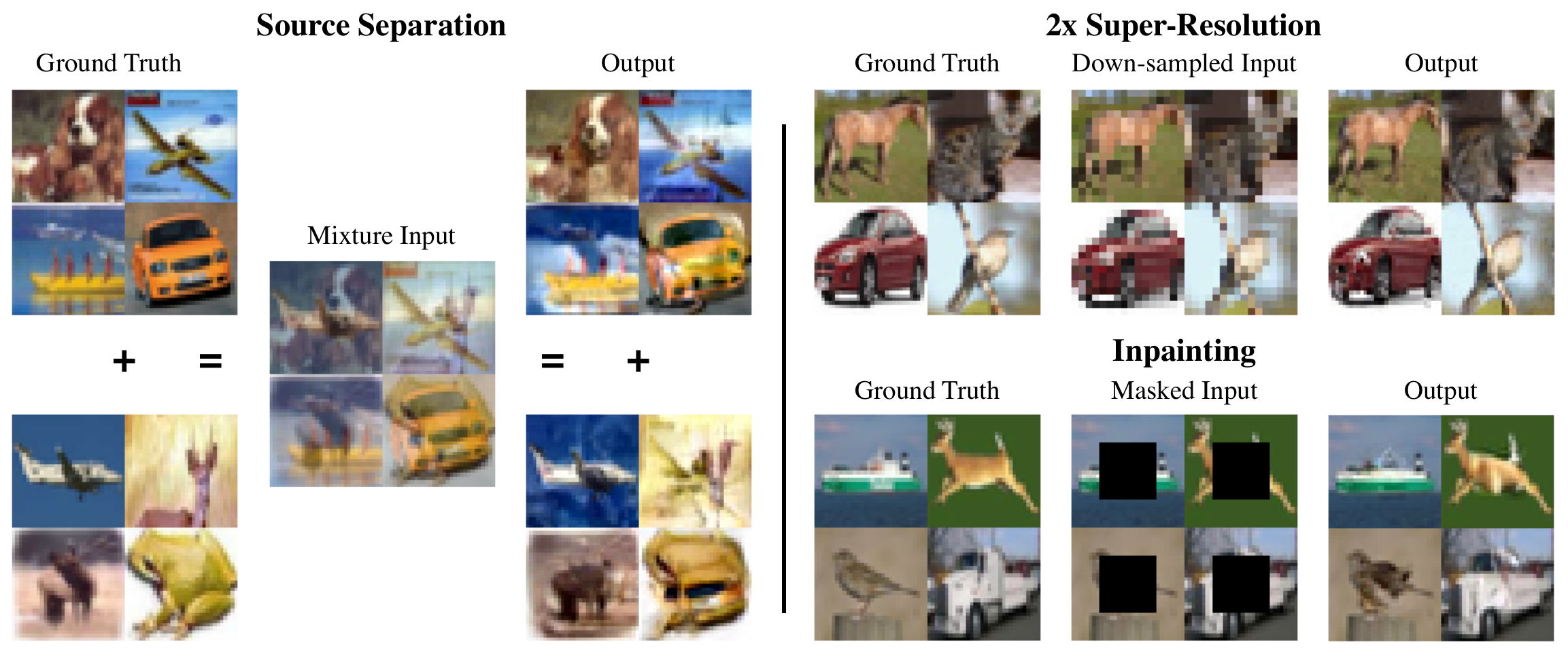}}
\caption{PnF sampling applied to visual source separation (Section~\ref{sec:source}) super-resolution (Section~\ref{sec:super}) and inpainting (Section~\ref{sec:inpaint}) using a PixelCNN++ prior over images trained on CIFAR-10. Ground-truth images in this figure are taken from the CIFAR-10 test set.}
\vspace{-5mm}
\label{fig:pixelcnn}
\end{center}
\end{figure*}

\subsection{Source Separation}\label{sec:source}

The single-channel source separation problem \citep{davies2007source} asks us to recover unobserved sources $\textbf{x}_1,\textbf{x}_2 \in \mathcal{X}^n = \mathbb{R}^{2\times n}$ given an observed mixture $\textbf{y} = g(\textbf{x}) = \textbf{x}_1 + \textbf{x}_2$. Like \citet{jayaram2020source}, we view source separation as a linear Bayesian inverse problem: recover $(\textbf{x}_1,\textbf{x}_2$) given $\textbf{y}$ and a prior $p(\textbf{x}_1,\textbf{x}_2) = p(\textbf{x}_1)p(\textbf{x}_2)$. We consider three variants of this task: (1) audio separation of mixtures of voice (VCTK test set) and piano (Supra Piano test set), (2) visual separation of mixtures of CIFAR-10 test set ``animal'' images and ``machine'' images, and (3) class-agnostic visual separation of mixtures of CIFAR-10 test set images.

We compare PnF audio separation to results using the Demucs \citep{defossez2019music} and Conv-Tasnet \citep{luo2019conv} source separation models. Both Demucs and Conv-Tasnet are supervised models, trained specifically for the source separation task, that learn to output source components given an input mixture. An advantage of PnF sampling is that it does not rely on pairs of source signals and mixes like these supervised methods. We train the supervised models on 10K mixtures of VCTK and Supra Piano samples and measure results on 1K test set mixtures using the standard Scale Invariant Signal-to-Distortion Ratio (SI-SDR) metric for audio source separation \citep{le2019sdr}. Results in Table~\ref{tab:audiosep} show that PnF sampling is competitive with these specialized source separation models. Qualitative comparisons are provided in the supplement. We do not compare results on the popular MusDB dataset \citep{MUSDB18} because this dataset has insufficient single-channel audio to train WaveNet generative models.

For CIFAR-10, we follow the experimental methodology described in \citet{jayaram2020source}. Table \ref{tab:cifar} shows that PixelCNN++ performs comparably to Glow as a prior, but underperforms NCSN. This makes sense, given the relative strength of NCSN as a prior over CIFAR-10 images in comparison to PixelCNN++ and Glow. Given the strong correlation between the quality of a generative model and the quality of separations using that model as a prior, we anticipate that more recent innovations in autoregressive image models based on transformers \citep{parmar2018image,child2019generating} will lead to stronger separation results once implementations of these models that match the results reported in these papers become public. Select qualitative image separation results are presented in Figure \ref{fig:pixelcnn}.

\begin{table}[h!]
\centering 
\caption{Quantitative results for audio source separation of mixtures of Supra piano and VCTK voice samples. Results are measured using SI-SDR (higher is better). }
\bigskip
\label{tab:audiosep}
  \begin{tabular}{l c c c}
  \hline
   Algorithm & \multicolumn{3}{c}{Test SI-SDR (dB)} \\
   \hline
   &  All & Piano & Voice \\
  \hline
  PnF (WaveNet) & 17.07 & 13.92 & 20.25 \\
  Conv-Tasnet & 17.48 & 20.02 & 15.50 \\
  Demucs & 14.18 & 16.67 & 12.75 \\

  \hline
  \end{tabular}
\end{table}

\begin{table}[h!]
\centering 
\caption{Quantitative results for visual sources separation on CIFAR-10. Results are measured using Inception Score / FID Score of 25,000 separations (50,000 separated images) of two overlapping CIFAR-10 images. In Class Split one image comes from the category of animals and other from the category of machines. NES results \citep{halperin2019neural} and BASIS results are as reported in \citet{jayaram2020source}. }
\bigskip
\label{tab:cifar}
  \begin{tabular}{l c c}
  \hline
  Algorithm & Inception Score & FID\\
  \hline
  Class Split\\
  \hline
  \hline
  NES & 5.29 $\pm$ 0.08  & 51.39\\
  BASIS (Glow) & 5.74 $\pm$ 0.05 & 40.21 \\
  \textbf{PnF (PixelCNN++)} & 5.86 $\pm$ 0.07 & 40.66 \\
  Average & 6.14 $\pm$ 0.11 & 39.49 \\
  BASIS (NCSN) & 7.83 $\pm$ 0.15 & 29.92\\
 \hline
  Class Agnostic\\
  \hline
  \hline
  BASIS (Glow) & 6.10 $\pm$  0.07 & 37.09 \\
  \textbf{PnF (PixelCNN++)} & 6.14 $\pm$ 0.15 & 37.89 \\
  Average & 7.18 $\pm$ 0.08 & 28.02 \\
  BASIS (NCSN) & 8.29 $\pm$ 0.16 & 22.12\\
 \hline
  \end{tabular}
\end{table}

\subsection{Super-Resolution}\label{sec:super}
The super-resolution problem asks us to recover unobserved data $\textbf{x}$ given a down-sampled observation $\textbf{y} = g(\textbf{x})$. For 1-dimensional (audio) super-resolution, $\textbf{x} \in \mathbb{R}^n$, $\textbf{y} \in \mathbb{R}^{n/r}$, and $\textbf{y}_i = \textbf{x}_{ri}$ \citep{kuleshov2017audio,eskimez2019adversarial}. For 2-dimensional (image) super-resolution $\textbf{x} \in \mathbb{R}^n = \mathbb{R}^{k\times k\times 3}$, $\textbf{y} \in \mathbb{R}^{k/r \times k/r\times 3}$, and $\textbf{y}_{i,j} = \textbf{x}_{ri,rj}$ \citep{dahl2017pixel,zhang2018residual}. Like source separation, super-resolution can be viewed as a Bayesian linear inverse problem, and we can recover solutions to this problem via PnF sampling. In the audio domain, the down-sampling operation $g(\textbf{x})$ can be interpreted as a low-pass filter.

\begin{table}[t]
	\centering
	\caption{\small Quantitative results for audio super-resolution at three different scales on the Supra piano and VCTK voice datasets. Results are measured using PSNR (higher is better). KEE refers to the method described in \citet{kuleshov2017audio}}
	\bigskip
	\label{tab:super}
	\small
	\begin{tabular}{c|ccc|ccc}
		\toprule
		& \multicolumn{3}{c}{Piano} & \multicolumn{3}{c}{Voice} \\  
		\midrule
		Ratio & Spline & KEE & PnF & Spline & KEE & PnF \\
		\midrule
		4x  & 23.07 & 22.25 & 29.78 & 15.8 & 16.04 & 15.47\\
		8x  &  13.58 & 15.79 & 23.49 & 10.7 & 11.15 & 10.03 \\
		16x  & 7.09 & 6.76 & 14.23 & 6.4 & 7.11 & 5.32 \\
		\bottomrule
	\end{tabular}
\end{table}

We measure audio super-resolution performance using peak signal-to-noise ratio (PSNR) and compare against a deep learning baseline \citep{kuleshov2017audio} as well as a simple cubic B-spline. Quantitative audio results are presented in Table~\ref{tab:super}, which show that we outperform these baselines on piano data and produce similar quality reconstructions on voice data. Qualitative audio samples are available in the supplement, where we also show examples of 32x super resolution---beyond the reported ability of existing methods.
Select qualitative visual results are presented in Figure~\ref{fig:pixelcnn}.

\subsection{Inpainting}\label{sec:inpaint}

Inpainting problems involve the recovery of unobserved data $\textbf{x}$ given a masked observation $g(\textbf{x}) = \textbf{m}\odot\textbf{x}$, where $\textbf{m} \in \{0,1\}^n$ \citep{adler2011audio,pathak2016context}. This family of problems includes completion tasks (finishing a sequence given a prime) pre-completion tasks (generating a prefix to a sequence) and outpainting tasks. Ancestral sampling can only be applied to completion tasks, whereas PnF sampling can be used to fill in any pattern of masked occlusions. Qualitative results for audio inpainting are available in the supplement. Select qualitative results for image inpainting are presented in Figure~\ref{fig:pixelcnn}.

%% file: src/conclusion.tex
\vspace{-1mm}
\section{Conclusion}
\label{sec:conclude}

In this paper we introduced PnF sampling, a parallelizable approach to sampling from autoregressive models that can be flexibly adapted to conditional sampling tasks. The flexibility of PnF sampling decouples the (unconditional) generative modeling problem from the details of specific conditional sampling tasks.
Using WaveNet models, we demonstrated a reduction in wall-clock sampling time using PnF sampling in comparison to ancestral sampling, as well as PnF's ability to solve a variety of practical audio processing problems: source separation, super-resolution, and inpainting.
We anticipate that PnF conditional sampling results will improve as developments in generative modeling empower us to incorporate stronger models as priors. 
More broadly, we are inspired by ongoing research in generative modeling that, coupled with PnF sampling, will continue to drive performance improvements for practical conditional image and audio restoration tasks.

%% file: src/appendix.tex
\section{PnF Sampling Details and Hyper-Parameters}

We broadly adopt the geometric annealing schedule and hyper-parameters of annealed Langevin dynamics introduced in \citet{song2019generative} and elaborated upon in \citet{song2020improved}. For both the PixelCNN++ and WaveNet models, we found that we needed additional intermediate noise levels to generate quality samples. We also found that good sample quality using these models required a smaller learning rate and mixing for more iterations than previous work \citep{song2019generative, jayaram2020source}. We speculate that the need for more levels of annealing and slower mixing could be a attributable to the autoregressive model parameterization, because also required a finer annealing and mixing schedule for the WaveNet models. Detailed hyper-parameters for the PixelCNN++ and WaveNet experiments are presented in Appendix \ref{sec:pcnn_details} and \ref{sec:wavenet_details} respectively.

Previous work makes the empirical observation that gradients of a noisy model $p_\sigma(\textbf{x})$ are inverse-proportional to the variance of the noise: $\mathbb{E} \|\nabla_\textbf{x}\log p_\sigma(\textbf{x}^{(t)})\|^2 \propto 1/\sigma^2$ \citep{song2019generative}. This motivates the choice of learning rate $\eta \propto \sigma^2$. The empirical scale of the gradients is conjectured in \citet{jayaram2020source} to be a consequence of ``severe non-smoothness of the noiseless model $p(\textbf{x})$.'' That work goes on to show that, if $p(\textbf{x})$ were a Dirac spike, then exact inverse-proportionality of the gradients would hold in expectation. For the discretized autoregressive models discussed in this work, the noiseless distribution $p(\textbf{x})$ is genuinely a mixture of Dirac spikes, and so the analysis in \citet{jayaram2020source} applies without caveats to these models and justifies the choice $\eta \propto \sigma^2$ (the precise constant of proportionality remains application dependent, and discussed in the experimental details sections).

\section{PixelCNN++ Experimental Details}\label{sec:pcnn_details}

Our visual sampling experiments are performed using a PixelCNN++ model $p(\textbf{x})$ trained on CIFAR-10. Specifically, we used a public implementation of PixelCNN++ written by Lucas Caccia, available at:

\url{https://github.com/pclucas14/pixel-cnn-pp}.

We used the pre-trained weights for this model shared by Lucas Caccia (at the link above) with a reported test-set log loss of 2.95 bits per dimension. For the models $p_\sigma(\textbf{x})$, we fine-tuned the pre-trained model for 10 epochs at each noise level $\sigma^2$.  We adopt the geometric annealing schedule proposed in \citet{song2019generative} for annealing $\sigma^2$, beginning at $\sigma_1 = 1.0$ and ending at $\sigma_{L} = 0.01$ using $L=19$ noise levels. This is double the number of noise levels used in previous work \citep{song2019generative,jayaram2020source}. We also found that sample quality improved using a smaller learning rate and mixing for more iterations than reported in previous work. For conditional sampling tasks, we set $\delta = 3e-06$ and $T=300$ in contrast to $\delta = 2e-05$ and $T=100$ used in previous work. In wall-clock time, we find that conditional PixelCNN++ sampling tasks require approximately 60 minutes to generate a batch of 16 samples using a 1080Ti GPU.

\section{WaveNet Experimental Details} \label{sec:wavenet_details}
Our audio sampling experiments are performed using a WaveNet model $p(\textbf{x})$ trained on both the VCTK and Supra Piano datasets. We used the public implementation of Wavenet  written by Ryuichi Yamamoto available at:

\url{https://github.com/r9y9/wavenet_vocoder}.

For all audio experiments, where data is encoded between $\{0..255\}$, we use $L=15$ noise levels geometrically spaced between $\sigma=175.9$ and $\sigma=0.15$. The same noise levels are also used for the sampling speed and quality results presented in Figures \ref{fig:likelihood} and \ref{fig:speed}. For all experiments, the number of Langevin steps per noise level is $T = 256$, except for Figure~\ref{fig:likelihood} where this parameter is varied to highlight changes in likelihood. The learning rate multiper $\delta = 0.05$ is used for all experiments. The Markov window $w$ is based on the underlying architecture. When training the fine-tuned noise models, all training hyperparameters are kept the same as the original WaveNet implementation. For the WaveNet implementation used in this paper, this is $6,139$ samples which is roughly 0.3 seconds at a 22kHz sample rate Please refer to the WaveNet paper or the public WaveNet implementation for training details.

As discussed in the WaveNet paper \cite{oord2016wavenet}, $8$-bit $\mu$-law encoding results in a higher fidelity representation of audio than $8$-bit linear encoding. For most experiments, the observation constraint $y=g(x)$ is still linear even under a $\mu$-law encoding of $x$. However, for source separation, the constraint $y = x_{1} + x_{2}$ is no longer linear under $\mu$-law encoding. Consequently, we use an $8$-bit linear encoding of $x$ for source separation experiments to avoid a change of variables calculation. To facilitate a fair comparison, all ground truths and baselines shown in the demos use the corresponding $\mu$-law or linear $8$-bit encoding.

In the source separation experiments, all mixtures were created with 1/2 gain on each source component. Due to the natural variation of loudness in the training datasets, we find that our model generalizes to mixtures without exactly 1/2 gain on each source. The real life source separation result on the project website shows that we can separate a mixture in the wild when we have no information about the relative loudness of each component.

\section{Additional PixelCNN++ Sampling Results}

\begin{figure*}[h!]
\begin{center}
\centerline{\includegraphics[width=\textwidth]{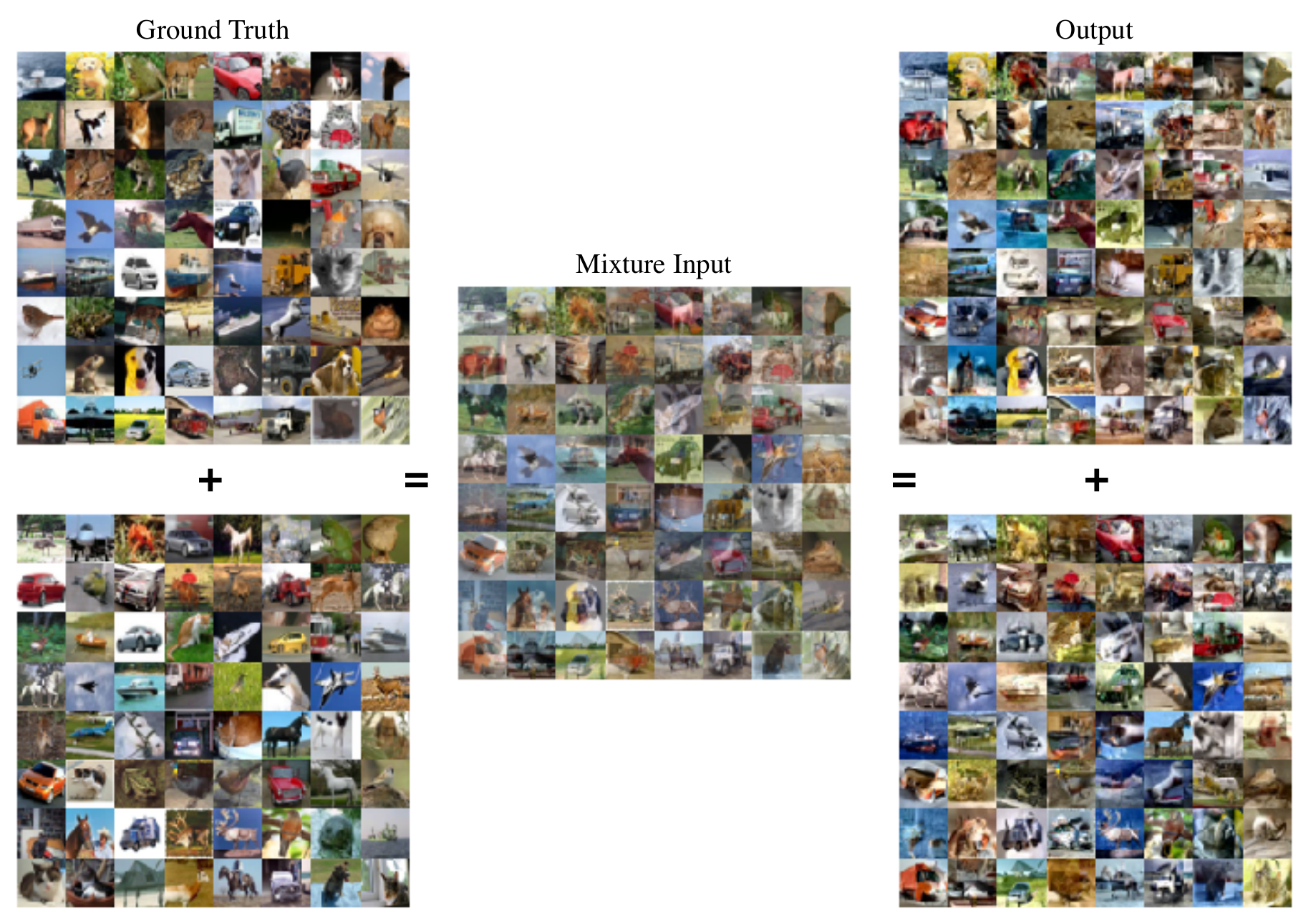}}
\caption{Additional uncurated results of PnF source separation (Section~\ref{sec:source}) for mixtures of CIFAR-10 test-set images using a PixelCNN++ prior trained on CIFAR-10.}
\label{fig:panel_sourcesep}
\end{center}
\end{figure*}

\begin{figure*}[h!]
\begin{center}
\centerline{\includegraphics[width=.9\textwidth]{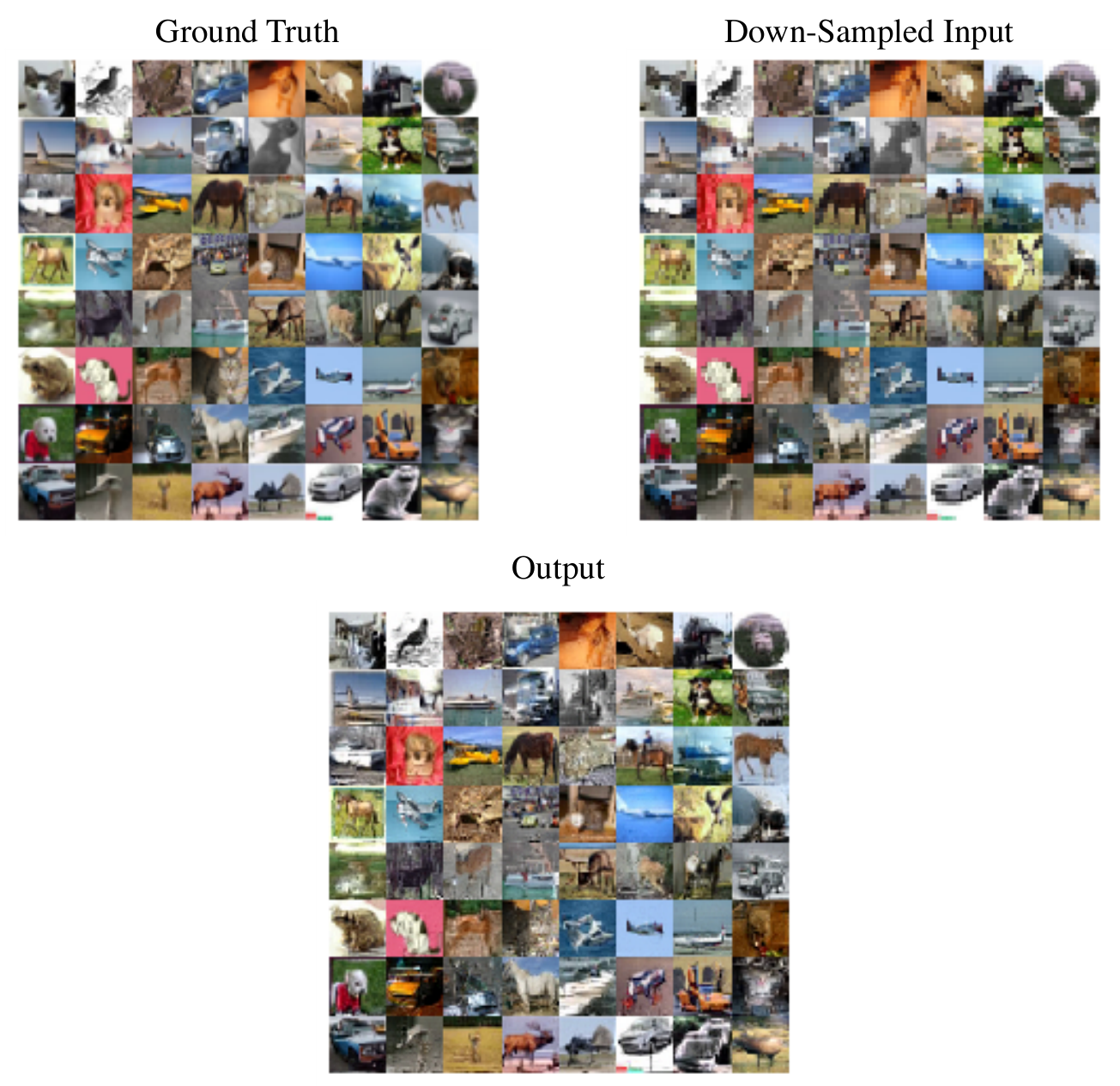}}
\caption{Additional uncurated results of PnF super-resolution (Section~\ref{sec:super}) applied to down-sampled CIFAR-10 test-set images using a PixelCNN++ prior trained on CIFAR-10.}
\label{fig:panel_super}
\end{center}
\end{figure*}

\begin{figure*}[h!]
\begin{center}
\centerline{\includegraphics[width=.9\textwidth]{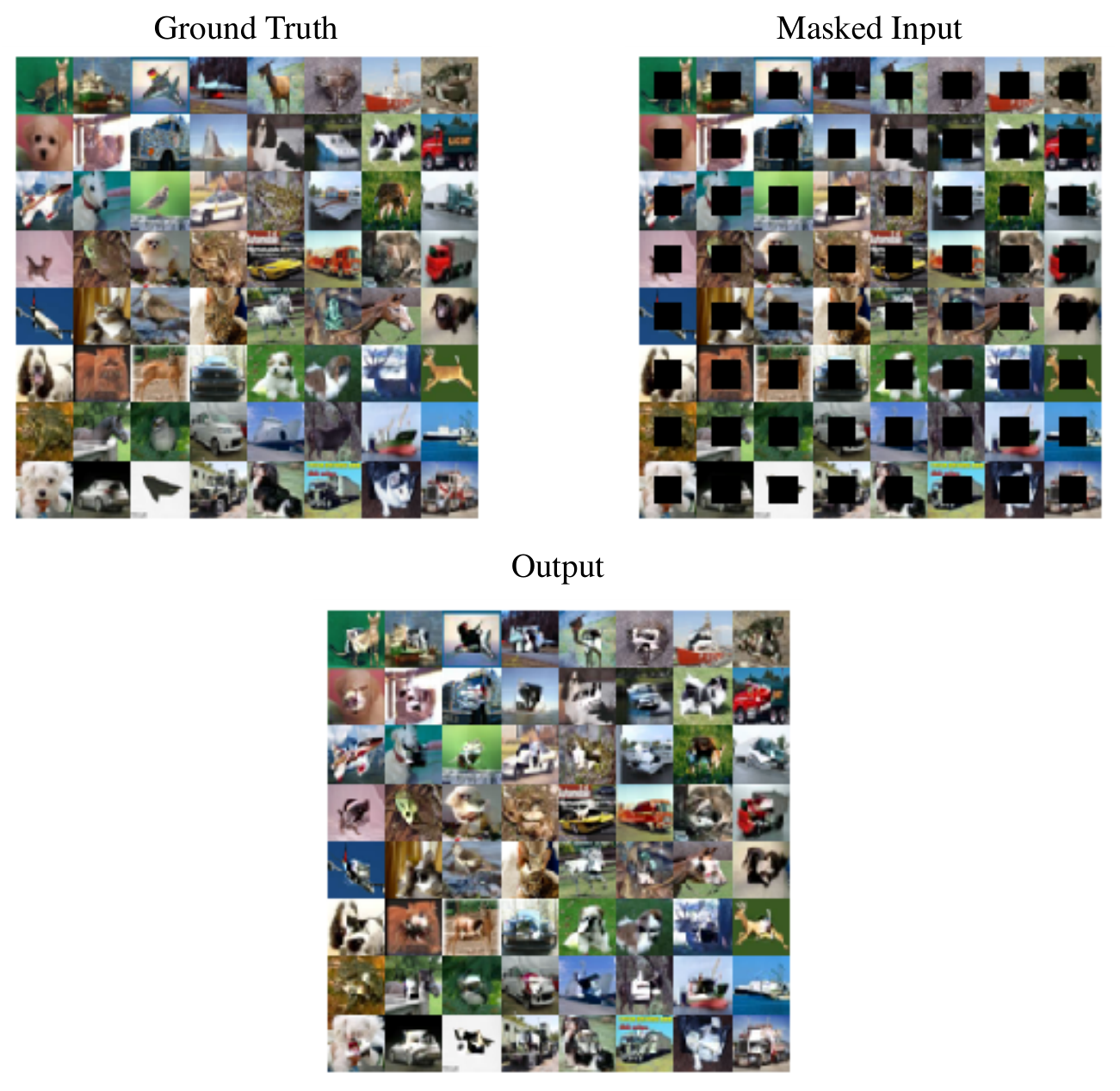}}
\caption{Additional uncurated results of PnF inpainting (Section~\ref{sec:inpaint}) applied to masked CIFAR-10 test-set images using a PixelCNN++ prior trained on CIFAR-10.}
\label{fig:panel_inpaint}
\end{center}
\end{figure*}